\documentclass[journal]{IEEEtran}
\ifCLASSINFOpdf
\else
\fi
\usepackage{amsmath}
\usepackage{amsfonts}
\usepackage{amssymb}
\usepackage[utf8]{inputenc}
\usepackage{graphicx}
\usepackage{caption}
\usepackage{subcaption}
\usepackage[export]{adjustbox}
\usepackage{algorithm}
\usepackage{algorithmic}

\usepackage{savesym}
\savesymbol{checkmark}
\usepackage{multirow}
\usepackage{graphicx}
\usepackage{dingbat}
\usepackage{pifont}
\usepackage{tabularray}
\usepackage{arydshln}
\graphicspath{{chapter5/}}

\begin{document}
%
\title{Single-Reference Text-to-Image Manipulation with Dual Contrastive Denoising Score}
%
%
%

\author{Syed~Muhammad~Israr,~
        Feng Zhao, 
\thanks{Syed Muhammad Israr and Feng Zhao are with the Department of Automation, School of Information Science and Technology, University of Science and Technology of China, Hefei 230026, China. (Email: misrarustc@mail.ustc.edu.cn)}} 

%
%

\markboth{Journal of \LaTeX\ Class Files,~Vol.~14, No.~8, August~2021}
{Shell \MakeLowercase{\textit{et al.}}: Bare Demo of IEEEtran.cls for IEEE Journals}
%



\maketitle

\begin{abstract}

Large-scale text-to-image generative models have shown remarkable ability to synthesize diverse and high-quality images. However, it is still challenging to directly apply these models for editing real images for two reasons. First, it is difficult for users to come up with a perfect text prompt that accurately describes every visual detail in the input image. Second, while existing models can introduce desirable changes in certain regions, they often dramatically alter the input content and introduce unexpected changes in unwanted regions. To address these challenges, we present Dual Contrastive Denoising Score, a simple yet powerful framework that leverages the rich generative prior of text-to-image diffusion models. Inspired by contrastive learning approaches for unpaired image-to-image translation, we introduce a straightforward dual contrastive loss within the proposed framework. Our approach utilizes the extensive spatial information from the intermediate representations of the self-attention layers in latent diffusion models without depending on auxiliary networks. Our method achieves both flexible content modification and structure preservation between input and output images, as well as zero-shot image-to-image translation. Through extensive experiments, we show that our approach outperforms existing methods in real image editing while maintaining the capability to directly utilize pretrained text-to-image diffusion models without further training.

\end{abstract}


%
\IEEEpeerreviewmaketitle

\section{Introduction}
\label{sec1}
The progress in Text-to-image (T2I) generative models is advancing rapidly and showing remarkable capability in generating a wide range of realistic images~\cite{chen2025conceptcraft,10193871,10288540,10375579}, using huge amounts of T2I datasets and compute resources. Emerging from Denoising Diffusion Probabilistic Models (DDPMs)~\cite{ho2020denoising} and score matching, Latent Diffusion Model (LDM)~\cite{rombach2022high} have demonstrated significant effectiveness in T2I tasks. Expanding on the advancements made by T2I models, several methods have been investigated to customize these models for text-guided image manipulation. The main goal of this process is apply the visual style from a reference source image while keeping the actual structure of the source image intact. This approach involved steering the generation process using gradient-based sampling techniques, or by directly training models that incorporate the source image as a condition. However, repurposing these approaches for customized image synthesis frequently encounter challenges in executing precise transfers, particularly in maintaining detailed features and ensuring that the transferred information remains accurate. First, images do not inherently include textual descriptions. Crafting a textual description is laborious and time-intensive, as each image conveys numerous textural features, lighting details, and shape nuances that may not be expressed in words. Second, despite the use of reference  and target text prompts (e.g., substituting the word dog with pig), current text-to-image models sometimes generate entirely novel output that does not adhere to the layout, shape, and object posture of the reference input image.

With the emergence of LDM, recent work~\cite{epstein2023diffusion,gal2022image,han2023svdiff,kumari2023multi,ruiz2023dreambooth} have started to utilize the capabilities of T2I models directly either by optimizing the distinct text embedding~\cite{radford2021learning} or the parameters of the diffusion model. Text-to-Image personalization~\cite{gal2022image,parmar2023zero,wang2024prior} and CatVersion~\cite{zhao2025catversion} inject new concepts into the latent space of diffusion model but fail to preserve the structural and shape details of the source image. Mask-based editing and inpainting methods~\cite{avrahami2023blended,nichol2022glide,ramesh2022hierarchical} require users to manually mask the area they want to edit. While these approaches can generate plausible results, the masking process is tedious and limits intuitive editing. Additionally, masking removes important structural information, hindering the ability to edit specific details like object textures. Score distillation sampling (SDS)~\cite{poole2022dreamfusion} demonstrates impressive performance in generating 3D assets by utilizing 2D diffusion model as prior and a 3D parametric generator i.e., NeRF~\cite{mildenhall2020nerf} but it often leads to semantically inconsistent and blurry outputs. Delta Denoising Score (DDS)~\cite{hertz2023delta} further extend SDS and introduce additional reference and target text-image pairs. It then queries the generative model using two pairs of images and texts. The target image is incrementally altered based on the difference between the two queries, resulting in a cleaner gradient direction however, it overlooks structural details in the source image. Contrastive Denoising Score (CDS)~\cite{nam2023contrastive} further improve DDS and introduce an additional contrastive loss to further preserve structural information in the final target image but it ignores certain parts of a detailed target text query.
Reflecting the textual details and preserving the structure of reference in the generated output is crucial in editing applications.
Expanding upon the effectiveness of these methods, similar concepts have been integrated into diffusion model, SpliceViT~\cite{tumanyan2022splicing} leverage features from ViT and ZeCon~\cite{yang2023zero} uses attention layers. Nonetheless, these adaptation methods have solely been employed in pixel domain, neglecting latent diffusion models, and  required supplementary encoder training, leading to inefficiencies.

To address the above mentioned issues, we propose an effective free from training method to leverage the internal representation of the latent diffusion model combined with dual contrasive learning objective (DualCUT). Specifically, we exploit the structural and style representations in the attention layers of Stable Diffusion (SD)~\cite{rombach2022high}. previously, the representations in the attention layers were only limited to transferring the style which lacks to preserve the structural features in the output results. We extract the key tokens from the attention layer of LDM and apply dual contrastive loss combined with Delta Denoising Score (DDS) to transfer the style feature in the target description and preserve the shape features in the reference source image. We visualize the internal representations of attention layers in Fig.~\ref{fig:feats}, which demonstrate that key tokens contain rich spatial cues for maintaining structural consistency.  Our main contributions are summarized as :
\begin{itemize}
	\item We introduce a novel approach that integrate dual contrastive loss with delta denoising score and attention layers representation, enabling structural and style consistency in the output images.
	\item Our work demonstrate that leveraging the hidden latent representation from attention layers allow for zero-shot image manipulation without requiring an additional network.
	\item Our proposed method demonstrates significantly improved performance, achieving a more effective balance between preserving structural details and transforming the content according to the target text prompt.
\end{itemize}

\section{Related Work}
\subsection{Text-to-Image Diffusion Models} Large-scale diffusion models~\cite{nichol2022glide,ramesh2022hierarchical,saharia2022photorealistic} have enabled the creation of photorealistic images using text-image datasets~\cite{saharia2022photorealistic,ramesh2021zero,yu2022scaling,gafni2022make}. While these methods offer some control over the generation process through text input, their capabilities are limited. Editing real images by altering words in the input sentence is unreliable as it often leads to significant and unpredictable changes in the image. Some recent approaches~\cite{nichol2022glide,avrahami2023blended} use additional masks to restrict the area where edits are applied. Palette~\cite{saharia2022palette},InstructPix2Pix~\cite{brooks2023instructpix2pix}, and PITI~\cite{wang2022pretraining} focus on developing conditional diffusion models specifically designed for image-to-image translation tasks. In contrast, our approach preserves the input structure without requiring any spatial masks. \newline
\subsection{Text-guided Editing with Diffusion Model} Text-guided image editing modifies an input image to match a text description. Numerous studies have investigated novel methods to achieve more precise control over the generative process~\cite{avrahami2023blended,bar2023multidiffusion,cao2023masactrl,li2023gligen,voynov2023sketch,zhang2023adding}, further unlocking their potential for a wide range of downstream editing task~\cite{avrahami2023blended,brooks2023instructpix2pix,choi2021ilvr}. Recent advancements have empowered users with greater flexibility in controlling both the creation and manipulation processes by incorporating spatial constraints defined by the users to pinpoint specific areas for editing~\cite{li2023gligen,avrahami2023spatext}. moreover, extensive research has underscored the effectiveness of internal representation of denoising networks, notably their attention blocks for enhanced image editing capabilities~\cite{parmar2023zero,patashnik2023localizing,liew2022magicmix}. Recently, Prompt-to-Prompt (P2P)~\cite{hertz2022prompt} introduce attention 
re-weighting to edit the specific attributes, Plug-and-Play (PnP)~\cite{tumanyan2023plug} inject self-attention maps to enable image-to-image translation. Null Textual Inversion~\cite{mokady2023null} and Proxedit~\cite{han2024proxedit} further extended these methods to real image editing that seamlessly blend with original image, pushing the boundaries of image manipulation. Although most image generation techniques are usually used in pre-trained models during the reverse process, Score Distillation Sampling (SDS)~\cite{poole2022dreamfusion} is an expanded technique that shows promise in 2D image and 3D object generating tasks. However, because SDS relies on the gradient of the difference between pure noise and the target text score prediction, it frequently generates over-smooth outputs. This constraint limits SDS ability to generate high-quality images to its maximum potential. To circumvent this, CDS~\cite{nam2023contrastive} proposed an alternative method to apply Contrastive loss on the output of self-attention layers in combination with DDS to improve the image quality. However, despite the impressive performance, CDS still lacks to maintain the alignment between the generated image and target text prompt. \newline
\subsection{Appearance Transfer with Diffusion Models} Appearance transfer can be categorized as special form of image translation. Classical Neural Transfer~\cite{gatys2016image,huang2017arbitrary} aims to transfer the global artistic style by iteratively matching the feature statistics. However, in this work, we focus on transferring the style between semantically related parts of the reference image and target image or text prompt. Generative Adversarial networks (GANs) based approaches~\cite{zhu2017unpaired,isola2017image,park2020contrastive} trained dedicated generators for each domain to transfer the structure and style features. However, these methods required separate generator and huge amount of paired and unpaired datasets.

Recently, Diffusion-based appearance transfer approaches~\cite{kwon2023diffusion,li2023gligen} have used the rich generative capabilities of large-scale diffusion models for style transfer without requiring further input or training. These methods typically utilize loss functions applied to the noisy latent codes. This guides the denoising process towards generating images that preserve the structural elements of the original image while allowing for adjustments to its appearance. These methods often struggle to achieve a fine-grained style transfer or limit the appearance transfer between objects that share similar characteristics, such as category, size, and shape. CDS~\cite{nam2023contrastive} took an effective step to combine DDS with contrastive loss on the output of self-attention layers to preserve the structural similarity. Despite CDS improved performance, it still lacks to align the output image with target text prompt. To address these limitations, we propose a novel method to effectively use the internal representation of attention blocks in Diffusion Models to effectively transfer the context in the target prompt and maintain the structural features of the reference source image.\newline

\section{Method}
\label{sec:method5}
Given a pair of reference image-prompt $(\mathcal{I}_{s}, \mathcal{P}_{s})$ and a target prompt $\mathcal{P}_{t}$, our goal is to generate an output image $\mathcal{I}_{t}$ that combines the structural elements of the reference $I_{s}$ with the description in the target prompt. To achieve this, we leverage a pretrained text-to-image diffusion model. specifically Stable Diffusion~\cite{rombach2022high}. Before delving into our approach, we provide a brief overview of the inner workings of diffusion models. We then introduce our proposed approach to faithfully adapt the structural attributes from the ref source image and style or semantic description from the target text prompt.
\subsection{Preliminaries}
\label{sec:prelim5}
\begin{figure}[t]
	\centering
	\includegraphics[width=\columnwidth]{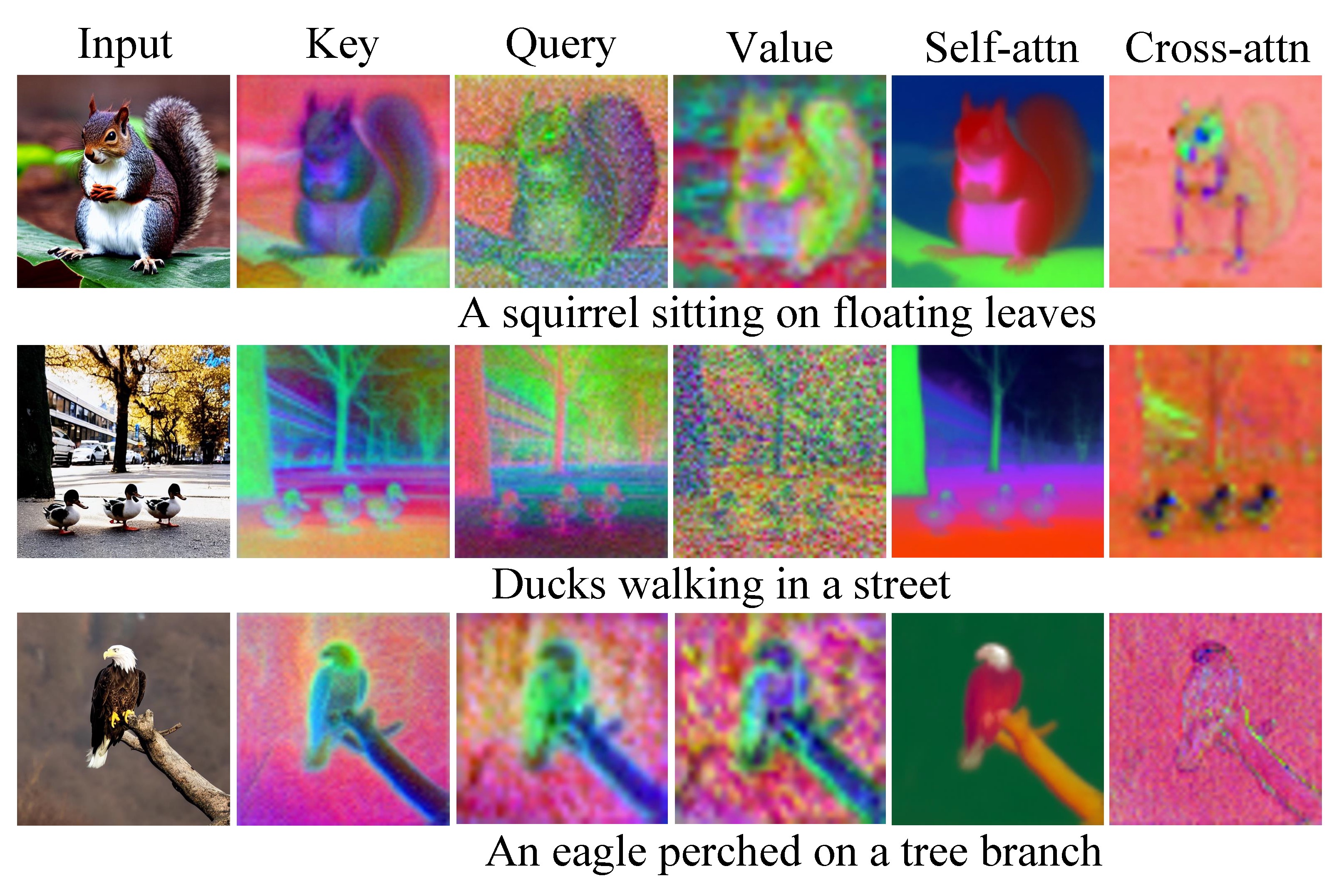}
	\caption{Visualization of the internal representation of the attention block. We visualized the key, query, value tokens, and the output of attention blocks. We observed that the key tokens of self-attention blocks contain rich spatial cues for applying Dual Contrastive Loss. Cross-attention contains style and contextual features aligned with the text prompt.  }
	\label{fig:feats}
\end{figure}

Diffusion Models have emerged as a powerful framework for generative modeling, particularly in text-to-image generation, through their iterative noise-to-data transformation. This section covers the essential concepts: the forward/reverse processes and attention mechanisms, which are fundamental to our DualCDS approach. DualCDS leverages pre-trained Stable Diffusion (SD)~\cite{rombach2022high} for training-free image editing while maintaining structure-semantics balance. \newline
\textbf{Diffusion Models:} generate images by denoising a random map $z_T$ through a noise prediction network $\epsilon_\theta$ (typically a U-Net) that estimates noise $\tilde{\epsilon}_t$ at step $t$. The forward process gradually adds noise to $x_0$ with variance $\beta_t \in(0,1)$ at $t \in[1, \ldots, T]$ as:
\begin{equation}
	\label{eq1}
	x_t=\sqrt{\bar{\alpha}_t} x_0+\sqrt{1-\bar{\alpha}_t} \epsilon,
\end{equation}
where $\epsilon \sim \mathcal{N}(0, I), \alpha_t=1-\beta_t$, and $\bar{\alpha}_t=\prod_{i=0}^t \alpha_i$. This process becomes a Markov chain that convert $x_0$ to nearly-isotropic Gaussian $x_T$. The noise removal also called the reverse process uses the network $\epsilon_\theta$  reconstructs real image $x_0$ by predicting the added noise at each step as:
\begin{equation}
	\label{eq2}
	x_{t-1}=\frac{1}{\sqrt{1-\beta_t}}\left(x_t-\frac{\beta_t}{\sqrt{1-\bar{\alpha}_t}} \epsilon_\theta\left(x_t, t\right)\right)+\sigma_t \epsilon,
\end{equation}
where $\sigma_t$ controls sampling randomness and $\epsilon_\theta(x_t,t)$ is the noise-predicting network. While DDPM performs these processes (Eq.~\ref{eq1}-\ref{eq2}) in pixel space, requiring intensive computation (typically T=1000 steps). DDIM adopts a non-Markovian approach in latent space via input encoding. The sampling process follows:

\begin{equation}
	\label{eq3}
	\begin{aligned}
		x_{t-1}= & \sqrt{\bar{\alpha}_{t-1}} \hat{x}_{0, t}\left(x_t\right) 
		& +\sqrt{1-\bar{\alpha}_{t-1}-\sigma_t^2} \epsilon_\theta\left(x_t, t\right)+\sigma_t^2 \epsilon.
	\end{aligned}
\end{equation}
Setting $\sigma_t=0$ yields deterministic denoising, enabling fewer steps (e.g., $T=50$) for faster sampling. In LDMs like Stable Diffusion, images are compressed to latent $z$ via a pretrained autoencoder, enabling efficient high-resolution synthesis (e.g., $512\times512$). The denoised output $\hat{x}_{0,t}(x_t)$ is:
\begin{equation}
	\hat{x}_{0, t}\left(x_t\right) = \frac{x_t-\sqrt{1-\bar{\alpha}_t} \epsilon_\theta\left(x_t, t\right)}{\sqrt{\bar{\alpha}_t}} .
\end{equation}
Diffusion models employ guidance mechanisms for conditional generation (e.g., T2I). Classifier guidance~\cite{dhariwal2021diffusion} uses a trained classifier $p_\phi(y|x_t,t)$ to condition on label $y$, adjusting the unconditional score $\epsilon_\theta(x_t,t)$ via classifier gradients: 

\begin{equation}
	\tilde{\epsilon}_\theta\left(x_t, t, y\right)=\epsilon_\theta\left(x_t, t\right)-\sqrt{1-\bar{\alpha}_t} \nabla_{x_t} \log p_\phi\left(y \mid x_t, t\right),
\end{equation}
where the gradient $\nabla_{x_t} \log p_\phi(y|x_t,t)$ steers $x_t$ toward label $y$ (e.g., "cat"). While effective, this requires training a classifier on noisy data, increasing complexity and limiting flexibility for arbitrary text prompts.
Classifier-free guidance~\cite{ho2021classifier} overcomes these issues by jointly learning conditional ($p(x|y)$) and unconditional ($p(x)$) distributions during training, randomly dropping $y$ (replacing it with $\emptyset$). At inference, the guided score interpolates between them:

\begin{equation}
	\nabla_x \log p_\omega(x \mid y)=(1-\omega) \nabla_x \log p(x)+\omega \nabla_x \log p(x \mid y),
\end{equation}
where $\omega$ balances conditional ($y$) and unconditional terms, with $y$ integrated via cross-attention in T2I models. \newline
\textbf{Attention Layers}: In diffusion UNet, attention layers update hidden representations using query, key, and value projections. In self-attention, intermediate features act as contextual information, linking image tokens from various regions for globally consistent structures. As shown in Fig.\ref{fig:feats}, $Q, K, V$, and attention features capture rich spatial cues such as object contours and layouts, with keys revealing detailed structures ideal for our dual contrastive loss. Cross-attention instead uses image features as queries and textual features, encoded via a CLIP text encoder, as keys and values, forming high-dimensional maps that link words to pixels~\cite{hertz2022prompt,tumanyan2023plug}. These maps guide expression and category alignment~\cite{hertz2022prompt,chefer2023attend} but may ignore structural details.
\begin{figure*}[t!]
	\centering
	\includegraphics[width=\textwidth]{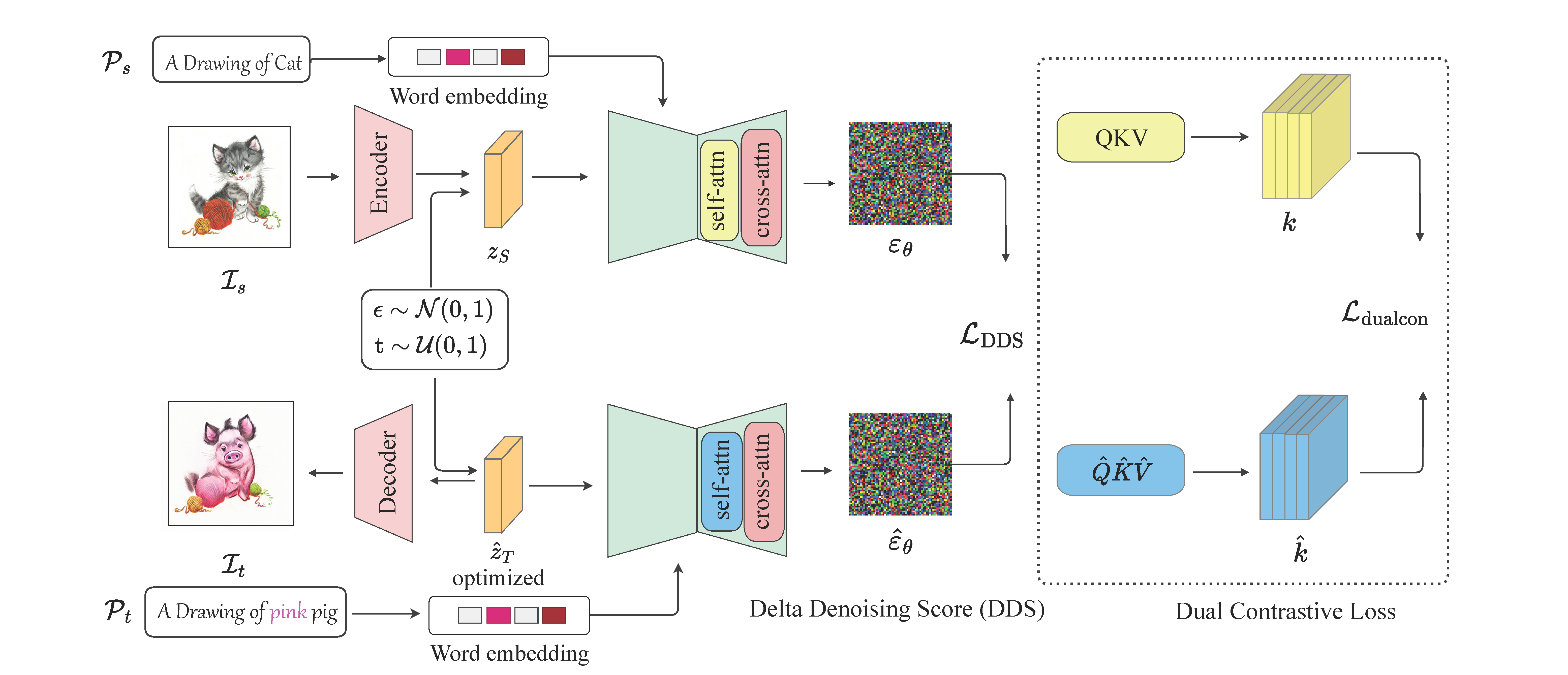}
	\caption{Framework of the proposed method. We extract the key features from the self-attention layer of the decoder part of the denoising Unet, as it contains structural details. We then apply dual contrastive loss $\mathcal{L}_{dualcon}$ on key vectors $k$,$\hat{k}$ of the reference and target images.}
	\label{framework}
\end{figure*}
\subsection{Dual Contrastive Learning for Structural Consistency}
Given an initial image $\mathcal{I}_s$ with textual description $\mathcal{P}_s$, our goal is to generate a target image $\mathcal{I}_t$ aligned with a new textual description $\mathcal{P}_t$, while preserving the structural integrity of $\mathcal{I}_s$. For example, an image $\mathcal{I}_s$ depicting a cat drawing linked to $\mathcal{P}_s$ should be translated into a pink pig drawing consistent with $\mathcal{P}_t$, as shown in Figure~\ref{framework}. This task is challenging, since diffusion-based methods without fine-tuning often fail to retain spatial cues~\cite{hertz2022prompt,tumanyan2023plug}. Simply applying $\mathcal{P}_t$ during generation produces images matching the prompt but with subjects that diverge from $\mathcal{I}_s$, even under fixed seeds~\cite{hertz2022prompt}.

To address this, we propose DualCDS, which merges the semantic layout from $\mathcal{P}_t$ with the content of $\mathcal{I}_s$. DualCDS leverages self-attention and cross-attention layers in the SD model to retrieve semantically consistent content from the source image, enabling the synthesis of $\mathcal{I}_t$ while preserving structure. Furthermore, inspired by Contrastive Unpaired Translation (CUT), we incorporate patch-wise contrastive learning into the denoising process, customized to exploit LDM internal representations.

Multiple attention blocks in the U-Net $\epsilon_\theta$ process the noisy latent $z_t$ at timestep $t$ during denoising. The hidden state $H$ in self-attention layer $l$ is learned as:
\begin{equation}
	\begin{split}
		H &= \operatorname{Softmax}\left(\frac{Q K^{\top}}{\sqrt{d}}\right) V, \\
		Q &= W_q H, \quad K = W_k H, \quad V = W_v H,
	\end{split}
\end{equation}
where $d$ represents feature dimension, $Q, K$, and $V$ are query, key, and value projections.  Inspired by SpliceViT, we extract key tokens $k$ (from $K$ ) and value tokens $v$ (from $V$ ), which encode rich spatial information.  We define negative as non-corresponding patches and positive pairs as corresponding patches between $k$ (from $\mathcal{I}_s$ ) and $\hat{k}$ (from $\mathcal{I}_t$), for $M_l$ patches at layer $l$.\newline
\textbf{Structural Consistency}: We use DualCDS to preserve structural consistency in the target images. The core concept of DualCDS is to leverage patch-wise contrastive learning within the feature space for unidirectional translation. The generated output should produce features with patches similar to corresponding patches from the input image while remaining distinct from other random patches, as illustrated in Figure~\ref{framework}. Unlike CUT, our DualCDS enforces an effective structural control useful in multimodal intricate image generation. Furthermore, unlike CUT-based approaches that require an auxiliary encoder, DualCDS directly computes the contrastive loss from latent representations in LDM, thereby streamlining the process while preserving structural consistency. Specifically, we extract the key tokens from the self-attention layers in the decoder part of LDM.
During the denoising process, at each timestep $t$, the noisy latent representation $z_t$ is processed by the denoising U-Net $\epsilon_\theta$, which is composed of multiple attention blocks. Within a self-attention layer $l$, the value tokens are denoted as $v$, which are multiplied by the attention matrix formed by query $q$ and key $k$ tokens. In cross-attention layers, the text prompt is combined with spatial image features via query and key tokens. During the reverse denoising process, we extract value tokens $v$ and $\hat{v}$ from the self-attention of the decoder part of the U-Net. We then select random patches from the value tokens, denoted as $m \in \{1, \ldots, M_l\}$, where $M_l$ is the total number of patches in the $l$-th layer. For each patch in the target key token $\hat{k}$, the corresponding patch in the source key token $k$ is treated as a positive sample, while all other patches are negatives. Our dual contrastive loss aims to maximize mutual information between positive pairs while minimizing it between negative pairs. We express $l_{pos}$ and $l_{neg}$ for positive and negative pairs as:

\begin{equation}
	\mathcal{L}_{\text {dualcon }} = \mathbb{E}_{\boldsymbol{k}}\left[\sum_l \sum_m \ell\left(\boldsymbol{k}, \hat{\boldsymbol{k}^m}, \hat{\boldsymbol{k}}^{M/m}\right)\right],
\end{equation}
where, $\ell(\cdot)$ represents the cross-entropy loss for positive patches:

\begin{equation}
	\ell_{pos}\left(\boldsymbol{k}, \boldsymbol{k}^{+}, \boldsymbol{k}^{-}\right)= 	-\log \left(\frac{e^{\left(k \cdot k^{+}/ \tau \right)}}{e^{\left(k \cdot k^{+}/ \tau \right)}+\sum e^{\left(k \cdot k^{-}/ \tau \right)}}\right),
\end{equation}
Similarly, for negative patches:

\begin{equation}
	\ell_{neg}\left(\boldsymbol{k}, \boldsymbol{k}^{-}, \boldsymbol{k}^{+}\right) = 	-\log \left(\frac{e^{-\left(k \cdot k^{-}/ \tau \right)}}{e^{-\left(k \cdot k^{-}/ \tau \right)}+\sum e^{-\left(k \cdot k^{+}/ \tau \right)}}\right).
\end{equation}
Consequently, $\ell_{pos}$ and $\ell_{neg}$ combined into total loss :
\begin{equation}
	\ell_{\text {dualcon }} = \lambda. \left( \ell_{pos} + \ell_{neg} \right),
\end{equation}
where $\lambda$ is a hyperparameter that controls the strength of the dual contrastive loss. For distant domains, a lower value up to 1 relaxes the structural constraint, yielding better performance, whereas in semantically similar domains, a higher value (up to 5) has shown satisfactory results.
\subsection{Style and Semantic Consistency}
While applying only dual contrastive loss, we noticed that the target output images were inconsistent with the textual prompts. To address this, we combined Delta Denoising Score (DDS)~\cite{hertz2023delta} with our DualCDS loss to achieve style-consistent results. DDS uses the denoising process inherent to diffusion models, where an image is gradually denoised from a random noise distribution to the final output. It introduces a comparative element by computing the denoising score for both the source image and the target prompt at various noise levels during the diffusion process. Mathematically, the denoising score for the source image is $\boldsymbol{\epsilon}_\phi\left(\boldsymbol{z}_s, \mathcal{P}_s, t\right)$, where $\boldsymbol{z}_s$ is the noisy version of the source image at timestep $t$, and $\phi$ represents the model parameters. Similarly, the denoising score for the target prompt is $\hat{\boldsymbol{\epsilon}}_\phi\left(\boldsymbol{z}_t(\theta), \mathcal{P}_t, t\right)$, where $\boldsymbol{z}_t(\theta)$ is the noisy target latent initialized from the source and optimized during training. The DDS loss is given as:
\begin{equation}
	\mathcal{L}_{\mathrm{DDS}}\left(\theta ; p_{t}\right)=\left\|\hat{\boldsymbol{\epsilon}}_\phi\left(\boldsymbol{z}_t(\theta), \mathcal{P}_t, t\right)-\boldsymbol{\epsilon}_\phi\left(\boldsymbol{z}_s, \mathcal{P}_s, t\right)\right\|^2.
\end{equation}
In this way, $\boldsymbol{z}_t(\theta)$ undergoes incremental updates in the direction of $\mathcal{L}_{\mathrm{DDS}}$ by minimizing the discrepancy between target and source domain. The total training loss is:
\begin{equation}
	\mathcal{L}_{\mathrm{total}} = \mathcal{L}_{\mathrm{dualcon}} + \mathcal{L}_{\mathrm{DDS}}.
\end{equation}
Optimization updates the target latent vectors as:
\begin{equation}
	z_t(\theta) \leftarrow z_t(\theta)-\eta \nabla_{z_t} \mathcal{L}_{\text {total }}.
\end{equation}
Combining $\mathcal{L}_{\text {dualcon }}$ and $\mathcal{L}_{\text {DDS}}$ harmonizes structural and semantic objectives. $\mathcal{L}_{\text{dualcon}}$ keep the spatial structure of $z_t(\theta)$ closely align with $z_t)$. $\mathcal{L}_{\text{DDS}}$ ensure the style and semantics description  $\mathcal{P}_t$ in the output images. Figure~\ref{main_results} shows the visual results of the proposed approach.
\section{Experiments}
\label{sec:exp5}
\subsubsection{Implementation Details} We build our method using Huggingface Stable Diffusion (SD) v1.5 and refer to the DDS implementation for diffusion-based image editing. Key features were extracted from self-attention layers, and PatchNCE loss was applied via dual contrastive learning across all upsampling and bottleneck blocks of the diffusion U-Net. For hyperparameters, we used patch sizes of 1 and 2 with $256$ patches in total. The weight of the contrastive loss $\lambda$ was set to 5.0 for semantically similar domains and 1.5 for distant domains. Default DDS settings were maintained for other parameters, including the number of optimization steps, choice of optimizer, and learning rate. All experiments were conducted on an NVIDIA RTX-3090 GPU, with each run taking approximately 4 minutes. \newline

\begin{figure*}[!ht]
	\centering
	\includegraphics[width=0.8\textwidth, height=0.8\textheight,keepaspectratio]{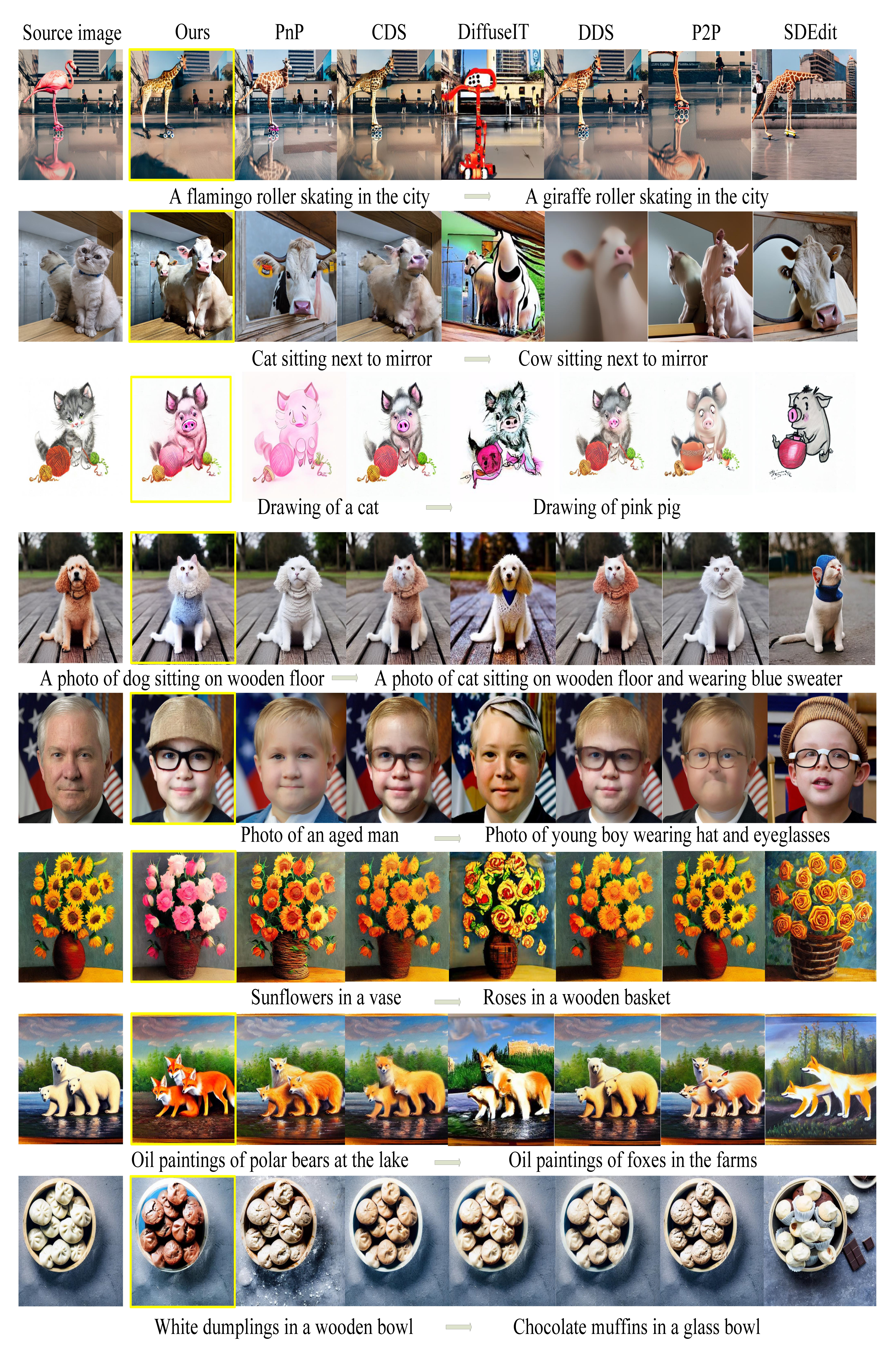}
	\caption{Visual comparison of our proposed method with the state-of-the-art methods.}
	\label{comparison}
\end{figure*}
\subsubsection{Baseline Models} To assess the effectiveness of our method, we conducted comparative experiments against several state-of-the-art techniques, including the original Delta Denoising Score (DDS). We use the Stable Diffusion version for SDEdit~\cite{meng2021sdedit}, Prompt-to-Prompt editing (P2P)~\cite{hertz2022prompt}, PnP~\cite{tumanyan2023plug}, CDS~\cite{nam2023contrastive}, and Mutual Self-attention Control (MasaCtrl)~\cite{cao2023masactrl} for image synthesis and editing. All experiments are performed using the officially recommended configurations.\newline
\begin{figure*}[t!]
	\centering
	\includegraphics[width=0.8\textwidth]{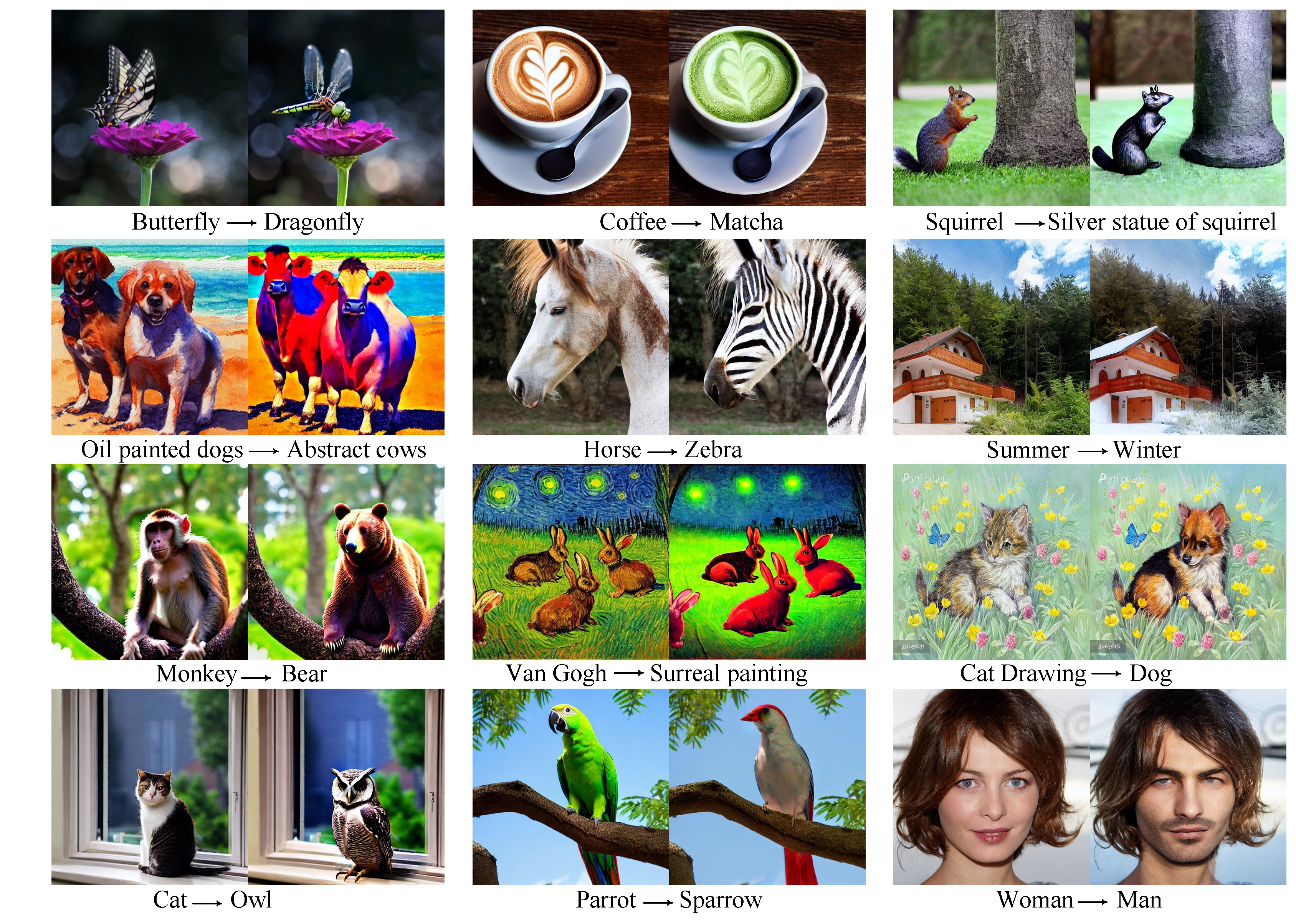}
	\caption{Qualitative Visual Results of the Proposed Method: Given a reference image and source and target text prompts, our method successfully generates target images that are structurally consistent with the reference image and precisely reflect the style and semantic description specified in the target prompt.}
	\label{main_results}
\end{figure*}
\subsubsection{Evaluation Matrices} For objective comparison, we use several quantitative evaluation metrics. CLIPScore~\cite{hessel2021clipscore} measure the alignment between generated image and target text prompt, where higher scores reflect better results. Structure Dist~\cite{tumanyan2022splicing} is a DINO based evaluation metric which measure the structural similarity between the reference and target image, where lower scores indicate better structural alignment. For background consistency, we use LPIPS, which measures similarity between background regions of the original and edited images. LPIPS is mainly applicable to foreground edits (e.g., cat to dog) and is not meaningful for global style changes (e.g., sketch to oil painting).
\subsection{Qualitative Results} 
Figure~\ref{main_results} shows visual results of the proposed method. Our approach, combining dual contrastive learning on internal attention representations with DDS, generates images that maintain content consistency. The synthesized images preserve the pose and structure of the source images, including both foreground and background elements, while accurately reflecting the target prompt $\mathcal{P}_t$, as illustrated in Figure~\ref{comparison}. 

Sampling-based techniques such as DiffuseIT~\cite{kwon2023diffusion} and SDEdit transform source attributes according to the target text but often suffer from severe structural deformations and undesirable artifacts. Attention-based methods such as Prompt-to-Prompt and MasaCtrl successfully reflect target prompt content but struggle to maintain structural consistency due to reliance on inversion quality and sensitivity to time-step selection for attention map modulation. The Plug-and-Play~\cite{tumanyan2023plug} baseline shows some improvement, generally preserving structure while aligning with text conditions, but fails to maintain consistency in challenging transformations (e.g., cat-to-cow) and often produces artifacts. CDS alleviates structural inconsistencies when combined with DDS but still fails to fully adhere to target prompts such as ``A pink pig sitting next to a mirror.''  

In contrast, our results demonstrate superior performance in balancing content preservation with prompt-guided transformation, a challenge that existing approaches have struggled to address.

\subsection{Quantitative Results}
\begin{table*}[t]
	\centering
	\caption{Quantitative comparison results of our proposed method. Our approach clearly outperforms current state-of-the-art methods in terms of text–image alignment, structural consistency, and background retention across various editing tasks.  }
	\label{table:ensemble}
	\scalebox{0.99}{
		\begin{tabular}{c|ccc|ccc}
			\hline \multirow{2}{*}{ Method } & \multicolumn{3}{c}{ cat2pig } & \multicolumn{3}{c}{ cat2cow } \\
			& CLIP Acc ( $\uparrow$ ) & Dist $(\downarrow)$ & LPIPS $(\downarrow)$ & CLIP Acc ( $\uparrow$ ) & Dist ( $\downarrow$ ) & LPIPS ( $\downarrow$ ) \\
			\hline SDEdit + word swap & $ 98. 6 \%$ & 0.062 & 0.127 & $97.8 \%$ & 0.059 & 0.133 \\
			DiffuseIT & $ 96.9 \%$ & 0.174 & 0.277 & $93.3 \%$ & 0.177 & 0.305 \\
			DDPM inv + P2P & $87.3 \%$ & 0.072 & 0.150 & $86.4 \%$ & 0.079 & 0.159 \\
			DDPM inv + PnP & $88.6 \%$ & 0.077 & 0.167 & $95.1 \%$ & 0.081 & 0.178 \\
			DDS & $ 99.5 \%$ & 0.032 & 0.088 & $ 9 9 . 6 \%$ & 0.042 & 0.117 \\
			CDS & $\mathbf{99.7} \%$ & $ 0.027$ & 0 .088 & $ 97.9 \%$ & $ 0 . 0 3 3$ & $ 0 . 1 1 3$ \\
			$\mathbf{Ours}$ & $\mathbf{99.7} \%$ & $ \mathbf{0.021} $ & $\mathbf{0 . 062}$ & $\mathbf{98.5} \%$ & $\mathbf{0 . 025}$ & $\mathbf{0 . 1 0 2}$ \\
			\hline
	\end{tabular}}
\end{table*}
We assess the generation quality of the proposed method using standard quantitative metrics. Following the standard protocols of Pix2Pix-Zero, we consider three tasks: (1) cat to pig, (2) cat to dog, and (3) cat to cow. We evaluate image editing using three criteria: edit success, structural consistency, and background preservation. These measure whether the desired changes are achieved, whether the overall structure of the original image is maintained, and whether irrelevant regions remain unaltered. We sample 250 images of cats from the LAION-5B~\cite{schuhmann2022laion} dataset using the source word cat. We then evaluate the target semantic description in the generated images using CLIPScore. We further assess structural alignment between the reference and target images using the DINO score. Table~\ref{table:ensemble} shows that our proposed method achieves higher CLIPScore and lower DINO structural distance, indicating better text–image alignment and editing consistency. We adopt the LPIPS distance to measure perceptual quality and background retention of the generated images. Our method results in lower LPIPS scores, demonstrating improved perceptual quality and background preservation.
\subsubsection{User Study} We conducted a human evaluation by presenting comparative results from seven models (including ours) to 30 participants. After reviewing the outputs, participants provided scores on a 1–10 scale. The evaluation covered three questions: (1) Text-to-image agreement: Do the images match the specified text? (2) Structural preservation: Does the target image retain the source structure? (3) Realism and quality: Are the generated images realistic and high quality? As shown in Table~\ref{table:usrstudy}, our method demonstrates superior performance across these criteria.  

\begin{table}[t]
	\centering
	\caption{Users study results. Our proposed method shows superior performance on all of the specified parameters. }	
	\label{table:usrstudy}
	\resizebox{0.97\columnwidth}{!}{
		\begin{tabular}{cccc}
			
			\hline \multirow{2}{*}{ Method } & \multicolumn{3}{c}{ Metric } \\
			\cline { 2 - 4 } & T2I agreement $(\uparrow)$ & Structure consistency $(\uparrow)$ & Realness $(\uparrow)$ \\
			\hline SDEdit + word swap & 7.54 & 5.80 & 6.86 \\
			DiffuseIT & 6.34 & 5.88 & 5.66 \\
			DDPM inv + P2P & 7.78 & 5.38 & 7.22 \\
			DDPM inv + PnP & 6.72 & 7.40 & 6.44 \\
			DDS & 8.12 & 8.11 & 7.28 \\
			CDS & 8.86 & 9.10 & 8.40 \\
			$\mathbf{Ours}$ & $\mathbf{9.34}$ & $\mathbf{9.42}$ & $\mathbf{9.12}$ \\
			\hline
	\end{tabular}}
\end{table}
\subsection{Ablation Study}
DualCDS components are ablated to better understand their roles in structure preservation, semantic alignment, and general editing quality.  Using 30 samples each from COCO and FFHQ datasets, experiments center on the cat-to-pig and house-to-church tasks, evaluated over 5 runs with measures including FID, CLIP Accuracy (CLIP Acc), Structure Distance (Struct Dist), and LPIPS.  With new visualizations and comparisons, this expanded study explores the theoretical and empirical responsibilities of every element and provides thorough insights into their interaction and performance effects.
\begin{table*}[t]
	\centering
	\caption{Ablation study demonstrating the impact of individual components in our proposed method. Starting with applying the contrastive loss on the value tensor, we then apply it to query and hidden states of the attention layer. Results illustrate that applying the contrastive loss on key effectively retains the structure preservation, background retention, and editing quality in the target images. }
	\label{table:ablation5}
	\scalebox{0.99}{
		\begin{tabular}{c|ccc|ccc}
			\hline \multirow{2}{*}{ Configuration } & \multicolumn{3}{c}{ cat2pig } & \multicolumn{3}{c}{ cat2cow } \\
			& CLIP Acc ( $\uparrow$ ) & Dist $(\downarrow)$ & LPIPS $(\downarrow)$ & CLIP Acc ( $\uparrow$ ) & Dist ( $\downarrow$ ) & LPIPS ( $\downarrow$ ) \\
			\hline w/ $\mathcal{L}_{DDS}$ & $85.22\%$ & 0.382 & 0.323 & $85.71\%$ & 0.384 & 0.436 \\ 
			$\mathcal{L}_{dcon}$ w/ hidden states & $86.2 \%$ & 0.215 & 0.189 & $86.7 \%$ & 0.324 & 0.288 \\
			$\mathcal{L}_{dcon}$ w/ enc. hidden states & $82.1 \%$ & 0.521 & 0.426 & $83.4 \%$ & 0.510 & 0.421 \\
			$\mathcal{L}_{dcon}$ w/ value & $ 91. 2 \%$ & 0.152 & 0.342 & $90.4 \%$ & 0.261 & 0.251 \\
			$\mathcal{L}_{dcon}$ w/query & $ 92.5 \%$ & 0.148 & 0.241 & $91.3 \%$ & 0.211 & 0.249 \\

			$\mathbf{\mathcal{L}_{dcon} w/ key (Ours)}$ & $\mathbf{99.7} \%$ & $ \mathbf{0.021} $ & $\mathbf{0 . 062}$ & $\mathbf{98.5} \%$ & $\mathbf{0 . 025}$ & $\mathbf{0 . 1 0 2}$ \\
			\hline
	\end{tabular}}
\end{table*}

\subsubsection{Effect of Dual Contrastive Loss $\mathcal{L}_{dualcon}$}
We performed an ablation study to evaluate the role of the proposed $\mathcal{L}{\text{dualcon}}$ loss in the DualCDS framework; results are shown in Figure~\ref{comparison}. This study isolates the impact of $\mathcal{L}{\text{dualcon}}$ by comparing outputs with and without it, emphasizing its role in balancing structural preservation and semantic alignment. Removing $\mathcal{L}_{\text{dualcon}}$ strips away contrastive learning, reducing the method to a vanilla Delta Denoising Score (DDS). In this baseline, structural details deteriorate even if content modulation aligns with the target prompt $\mathcal{P}t$. For instance, turning a "cat drawing" into a "pink pig drawing" without the loss leads to distorted limbs or omission of attributes like "pink." In contrast, including $\mathcal{L}{\text{dualcon}}$ preserves layout and pose, keeping $\mathcal{I}_t$ consistent with $\mathcal{I}_s$ while reflecting stylistic changes.

Quantitative and qualitative results confirm this. LPIPS increases significantly, and structural distance rises when $\mathcal{L}{\text{dualcon}}$ is removed, indicating perceptual degradation and layout drift—despite CLIP accuracy remaining high. This suggests DDS prioritizes semantic alignment over structural integrity. For example, in transforming "sunflowers in a vase" to "roses in a wooden basket," outputs without DualCut loss capture the basket but ignore the rose color. By contrast, $\mathcal{L}{\text{dualcon}}$ enables our method to maintain coherence in both structure and content, effectively anchoring the layout while aligning with the target prompt.

\subsubsection{Feature Space Selection for DualCUT Loss}
\begin{figure}[t!]
	\centering
	\includegraphics[width=\linewidth]{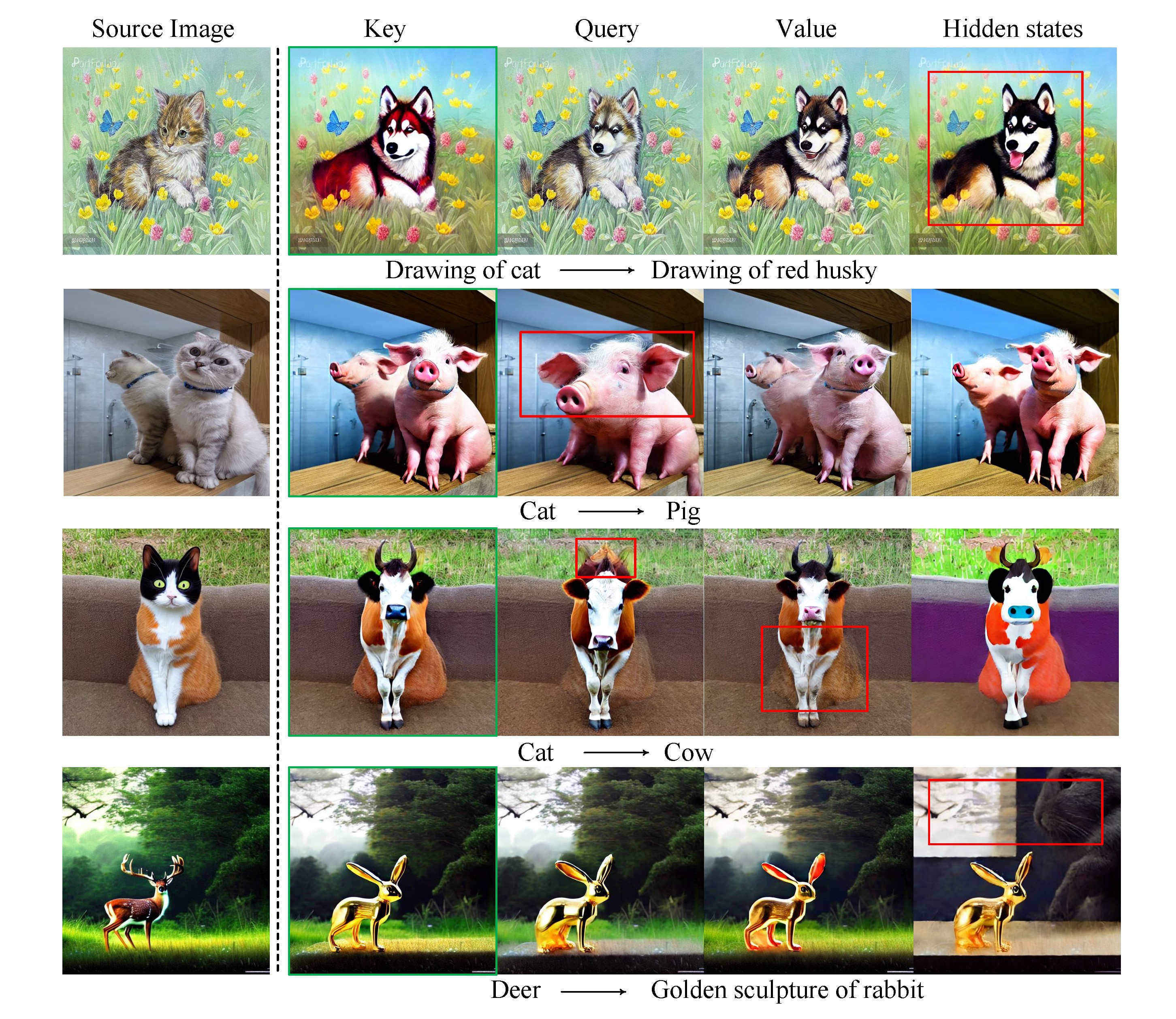}
	\caption{Feature selection for applying dual contrastive loss $\mathcal{L}_{dualcon}$. The visual results demonstrate that applying loss on key tokens effectively preserves the structural details of the reference source image.}
	\label{loss_loc}
\end{figure} 
We conducted a detailed analysis to evaluate different feature extraction layers for computing $\mathcal{L}{\text{dualcon}}$ in DualCDS, as shown in Figure~\ref{loss_loc}. Specifically, we examined applying the loss to (i) the self-attention layer output, (ii) the cross-attention hidden state, and (iii) the query, key, and value vectors from self-attention. Each setup explores the trade-off between semantic alignment and structural preservation during editing. Applying $\mathcal{L}{\text{dualcon}}$ to the full self-attention output overly constrains the network, preserving structure but limiting semantic adaptation. For instance, turning a "cat next to mirror" into a "pig next to mirror" retains the pose and background but fails to depict pig-specific features. In contrast, applying the loss to cross-attention hidden states emphasizes prompt alignment but weakens spatial consistency, producing outputs with altered postures or distorted backgrounds that diverge from the reference $\mathcal{I}_s$.

The most effective setup applies the loss to key vectors from self-attention layers. This captures spatial cues like object boundaries and positions, preserving layout while enabling accurate semantic changes. For example, in transforming "drawing of cat" to "red husky" (Figure~\ref{loss_loc}), the edited image retains posture and context while integrating new features. Key vectors provide a precise spatial representation suited to the patch-level contrastive goal of $\mathcal{L}_{\text{dualcon}}$. Visual results in Figure~\ref{loss_loc} confirm this setup best maintains structural integrity and semantic fidelity. These findings support our design choice and highlight the importance of feature selection in maximizing the effectiveness of contrastive learning for high-quality, structure-aware image editing.

\begin{figure}[t!]
	\centering
	\includegraphics[width=\linewidth]{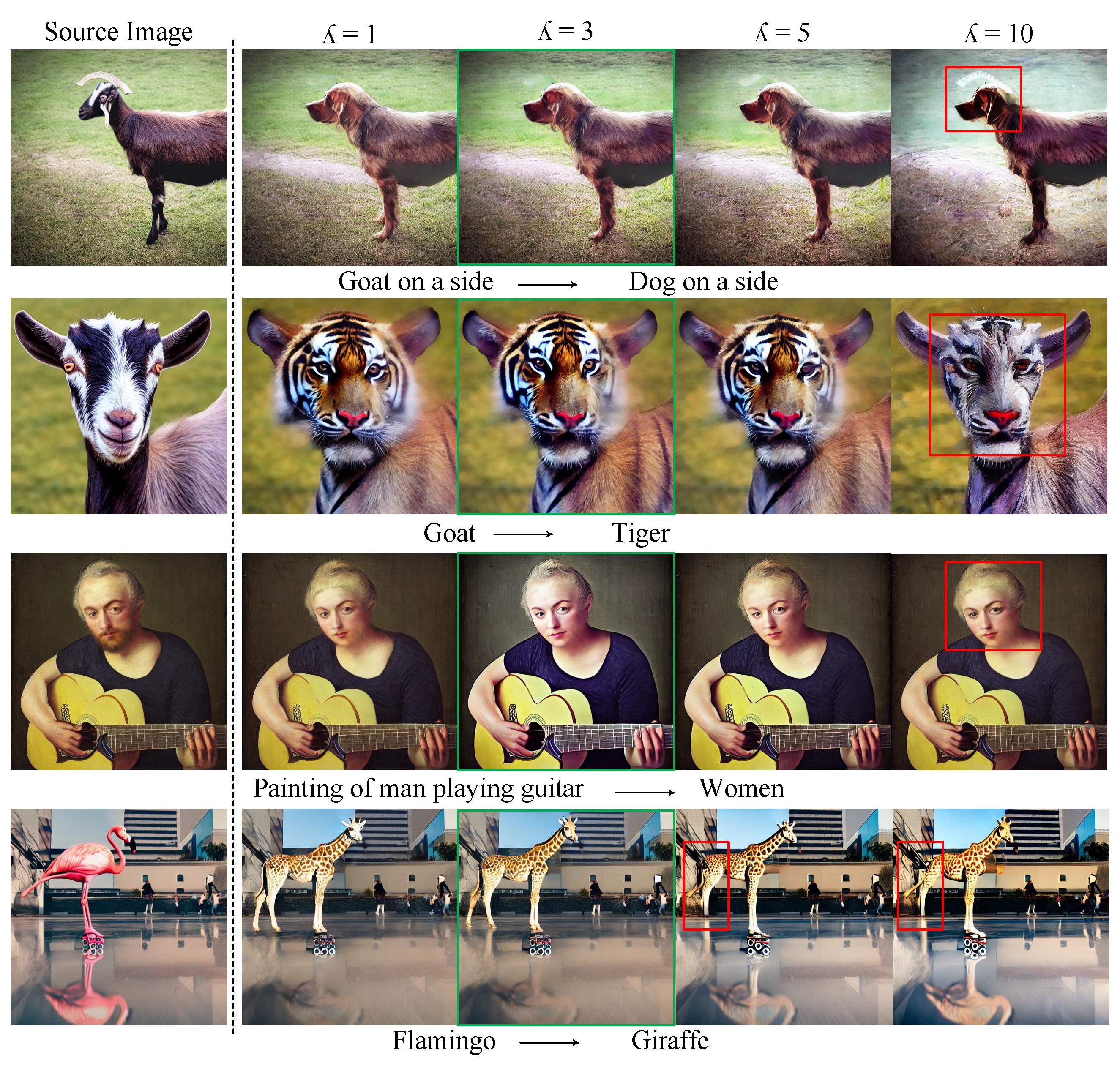}
	\caption{Visual results of various loss coefficient values $\lambda$. We demonstrate that a lower value relaxes the structural constraint, while a higher value overly restricts it, resulting in poor image quality. We found that maintaining $\lambda=3$ as the default yields better performance.  }
	\label{loss_value}
\end{figure}

\subsubsection{Impact of DualCUT Loss Coefficient on Edibility and Consistency  }
We conducted comprehensive experiments by varying the DualCUT loss coefficient to examine its effect on output quality and editing behavior. This analysis aimed to understand how the contrastive loss weight influences the balance between semantic editability and structural consistency in the output image $\mathcal{I}_t$. Results show a strong correlation between editing characteristics and the coefficient value. Higher weights (e.g., $\geq5$) emphasize structure preservation, retaining both foreground and background elements from the reference image $\mathcal{I}_s$. For example, in converting a "goat on a side" to a "dog on a side" (Figure~\ref{loss_value}), high coefficients maintain posture and contextual details while enabling semantic change. In contrast, smaller weights (e.g., $\le1$) prioritize prompt alignment, often at the cost of structural fidelity. As seen in the "flamingo" to "giraffe" case, low weights lead to posture shifts or missing limbs, compromising spatial coherence with $\mathcal{I}_s$.

Our findings indicate that $\lambda = 3$ offers an optimal trade-off, preserving structure while allowing clear semantic incorporation such as giraffe-specific traits. Values beyond 3 (e.g., 7 or 10) overly constrain changes, reducing expressiveness, while smaller values (e.g., 0.5) introduce structural drift and instability. As shown in Figure~\ref{loss_value}, the coefficient acts as a key control for managing structural retention and flexibility in semantic updates. This parametric study highlights its role as a vital hyperparameter in DualCUT, enabling users to fine-tune the editing process depending on task requirements, favoring structural precision in subtle edits or permitting broader changes when semantic freedom is prioritized.
\subsubsection{Size of Patch Window}
\begin{figure}[t!]
	\centering
	\includegraphics[width=\linewidth]{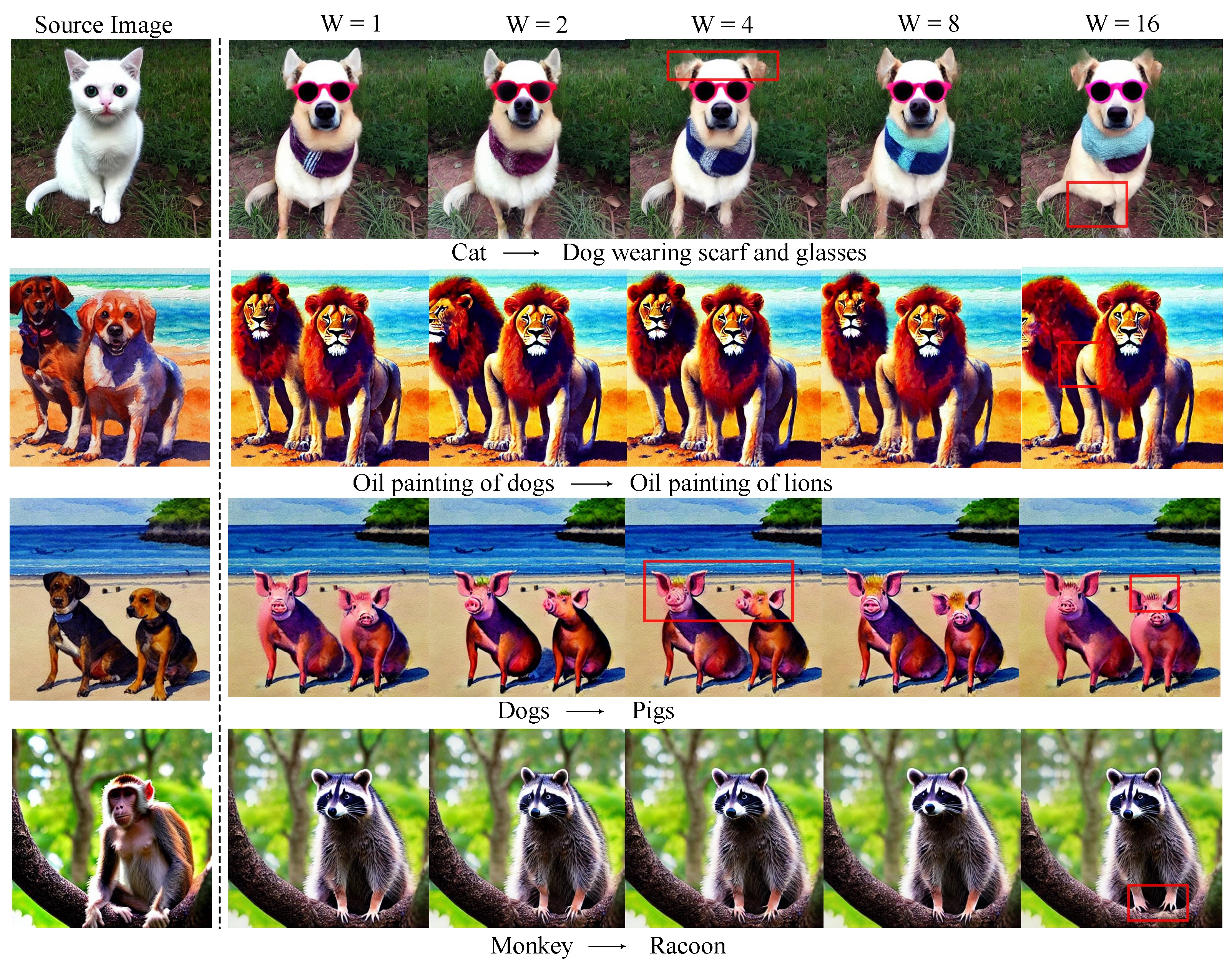}
	\caption{Impact of patch window size $w$. We ablate the size of the patch window for applying dual contrastive loss. Our experiments demonstrated that a lower value of $w$ as 1 preserves more finer details and a higher value up to $w=16$ focuses more on global features. We found that using setting values of 1 and 2 leads to optimal results.}
	\label{p_size}
\end{figure}

To evaluate the impact of patch size on DualCUT performance within DualCDS, we examined its role in preserving structure and embedding semantic cues from the target prompt. As shown in Figure~\ref{p_size}, larger patches (e.g., $4\times4$) reduce structural coherence and blur fine details by aggregating over broad regions. For example, transforming a “cat” into a “dog wearing scarf and glasses” with large patches leads to distorted facial features and loss of semantic elements like the dog’s snout or accessories. This is due to reduced spatial precision in the latent space. Smaller patches ($1\times1$, $2\times2$), on the other hand, preserve posture, contours, and background elements more effectively. In Figure~\ref{p_size}, a $2\times2$ setting retains the cat's pose and scarf while incorporating the new dog-specific features, reflecting improved semantic accuracy and spatial consistency.

Based on these results, $1\times1$ and $2\times2$ patches are chosen as optimal for DualCUT. They balance structural integrity with semantic alignment more effectively than larger patches. Figure~\ref{p_size} visually confirms this, showing that smaller patches yield sharper, more coherent edits. This underscores the importance of patch size in enhancing DualCUT’s ability to guide high-quality, training-free image manipulation.

\subsubsection{Number of Image Patches}
We conducted ablation studies to evaluate how the number of image patches influences the balance between structural preservation and semantic modification in the DualCDS framework. Increasing the patch count improves the model’s ability to retain structural details, including facial features, head positions, and posture, while embedding target semantics. For example, transforming a “fox sitting in a basket” into a “poodle wearing blue sweater and sitting in a basket” benefits from higher patch counts, ensuring features like ear alignment, limb curvature, and basket placement remain intact. With more patches, the image better reflects the reference structure while adapting to the target prompt.

\begin{figure}[t!]
	\centering
	\includegraphics[width=\linewidth]{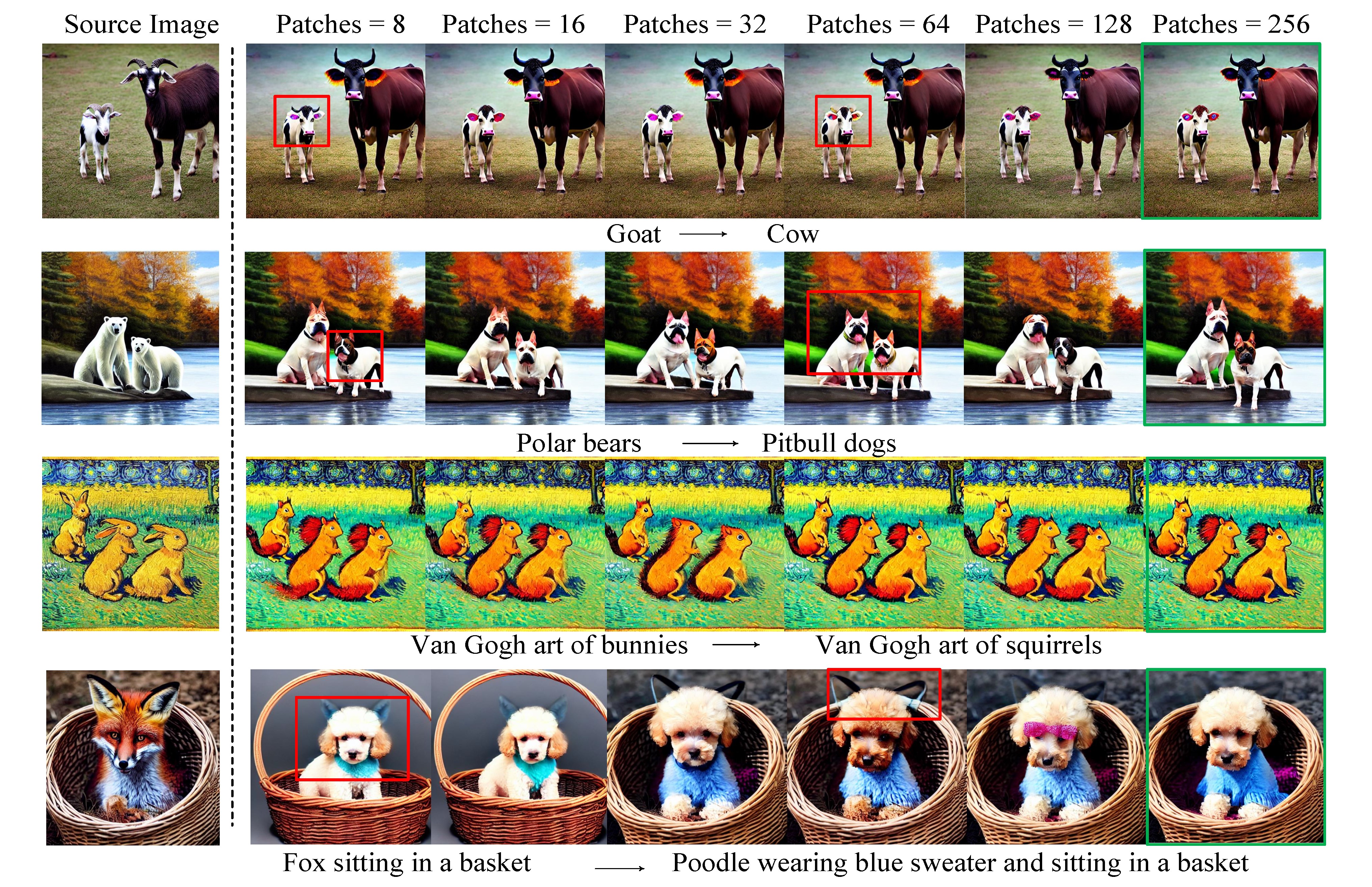}
	\caption{Visualizing the Impact of Number of Patches (Tokens): We have experimented with different numbers of patches as context length for $\mathcal{L}_{dualcon}$. Our results demonstrated that using a lower number of image patches leads to incomplete manipulation results. Conversely, using a higher number of patches yields fine-grained editing outcomes.}
	\label{n_patches}
\end{figure}

We varied the patch number from 8 to 512 and examined how this affected output quality. At lower counts (8 to 16), the model’s performance dropped, producing incomplete edits and losing structural coherence. The reference posture would shift, leading to a poodle with mismatched stance or blurred edges. In contrast, more patches preserved spatial details and improved alignment with both structure and semantics. However, increasing patches beyond 256 gave limited returns. The output quality improved slightly, but the added memory and computation were not justified. Comparing 256 with 512 patches showed minimal difference despite higher resource use. Based on this, we select 256 patches as the optimal configuration, offering a practical trade-off between editing quality and efficiency. Figure~\ref{n_patches} summarizes these findings and shows how patch quantity influences contrastive learning in our method.
\begin{figure}[t!]
	\centering
	\includegraphics[width=\linewidth]{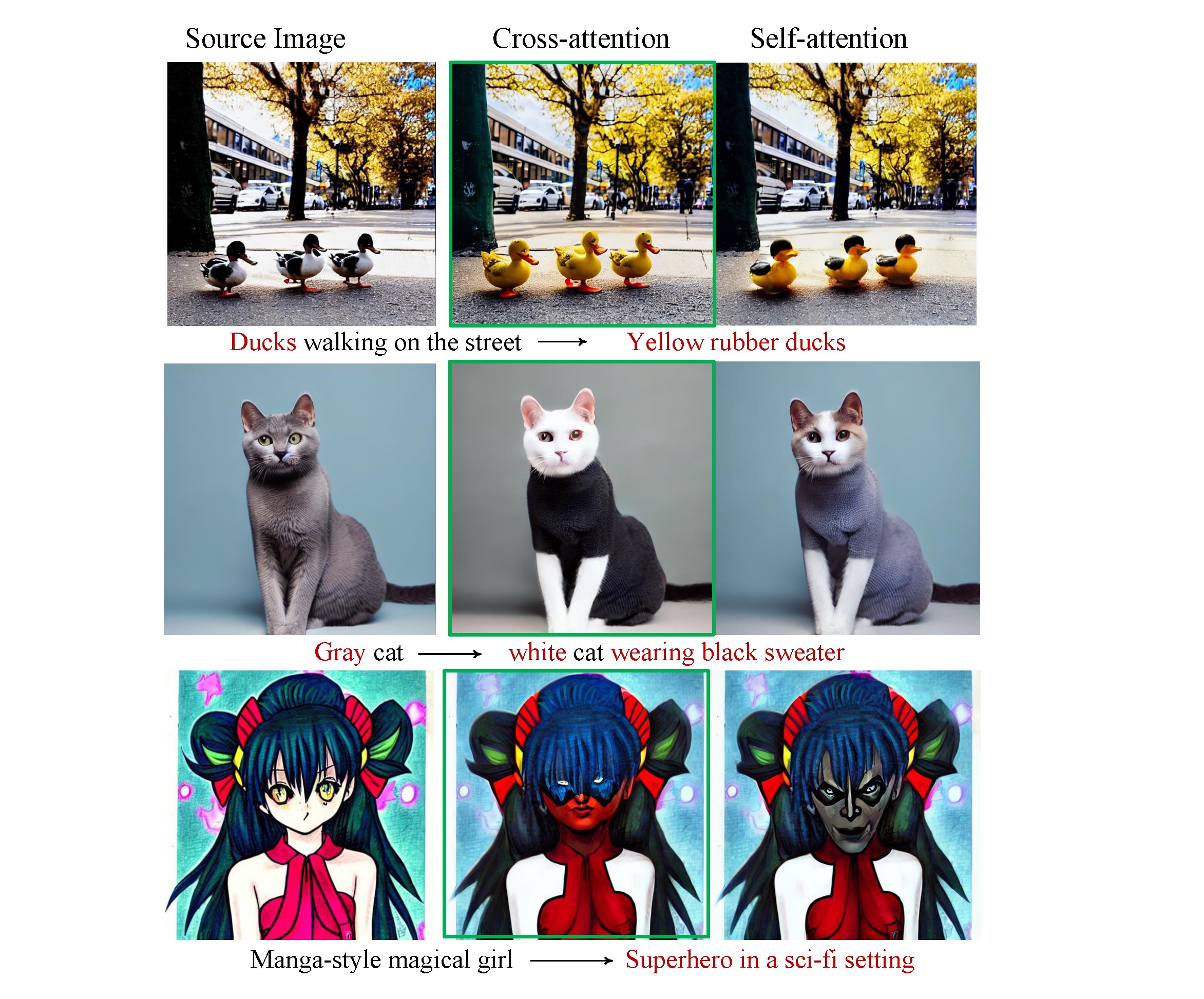}
	\caption{Choosing Parameters for Optimization: We experimented with updating parameters in self and cross-attention layers and observed that applying loss to key tokens of self-attention layers while updating the parameters of cross-attention layers leads to coherent editing results.}
	\label{loss_optim}
\end{figure} 
\subsubsection{parameter selection for optimization}
We perform an ablation study on the self-attention and cross-attention layers of the Stable Diffusion U-Net to evaluate their impact on the DualCDS technique. The results reveal distinct roles for each layer type in achieving structural preservation and semantic alignment. Specifically, modifying cross-attention parameters significantly enhances text-to-image coherence, allowing the generated image $\mathcal{I}_t$ to better reflect the stylistic and semantic features of the target prompt $\mathcal{P}_t$. For instance, when converting a “gray cat” into a “white cat wearing a black sweater,” tuning the cross-attention layers ensures accurate inclusion of color, clothing, and pose aligned with the prompt. In contrast, optimizing self-attention layers aids in preserving structural details such as pose and spatial layout but struggles to capture style and semantics. As shown in Figure~\ref{loss_optim}, changes to cross-attention produce outputs more faithful to the prompt, balancing structure with semantics, as illustrated in the green frames. These findings demonstrate the complementary nature of the attention layers and support our strategy of prioritizing cross-attention to align structure and style in the output image.
\section{Conclusion}
In this work, we presented DualCDS, a novel text-to-image editing method utilizing dual contrastive learning with latent diffusion models. Our method addresses two critical challenges: preserving desired edits while maintaining structural consistency and accurate alignment with target text prompts. We leverage the spatial information in self-attention layers to construct features input into our dual contrastive loss. This preserves structural details of the source image while transferring semantic changes from the target prompt. Experimental results across editing tasks show DualCDS outperforms prior methods in both editing quality and structural coherence. Ablation studies demonstrate the effectiveness of each component, especially the role of contrastive loss on key vectors in self-attention layers and its synergy with delta denoising score. Our method achieves these results without additional training or auxiliary networks, making it practical for real-world use. Despite challenges in balancing semantic edits and structural integrity, DualCDS shows strong performance across text-to-image tasks. Future work may extend this to more complex scenarios including video and 3D synthesis. The results illustrate that leveraging internal representations of pre-trained diffusion models with targeted loss functions enables more controllable and effective image editing.


%

\appendices

\section*{Acknowledgment}

The authors would like to thank...

\ifCLASSOPTIONcaptionsoff
  \newpage
\fi



\bibliographystyle{IEEEtran}
\bibliography{IEEEexample}
\end{document}